\pdfoutput=1

\documentclass[11pt]{article}

\usepackage{EMNLP2023}

\usepackage{times}
\usepackage{latexsym}

\usepackage[T1]{fontenc}

\usepackage[utf8]{inputenc}

\usepackage{microtype}

\usepackage{inconsolata}

\usepackage{graphicx}
\graphicspath{ {./diagrams/} }

\usepackage {algorithm}
\usepackage {algpseudocode}
\usepackage{amsmath}
\usepackage{dblfloatfix}

\usepackage{pgfplots}

\title{InsightNet : Structured Insight Mining from Customer Feedback}

\author{Sandeep Sricharan Mukku\,, Manan Soni\,, Jitenkumar Rana\,, Chetan Aggarwal\,, \\ \vspace{0.5cm}
    {\bf Promod Yenigalla\,, Rashmi Patange}\and{\bf Shyam Mohan} \\
        Amazon \\ 
        \textit{\texttt{\{smukku, sonmanav, jitenkra, caggar, promy, rpatang, shyaamm\}@amazon.com}} }

\pgfplotsset{compat=1.18}
\begin{document}
\maketitle

\begin{abstract}
We propose InsightNet, a novel approach for the automated extraction of structured insights from customer reviews. 
Our end-to-end machine learning framework is designed to overcome the limitations of current solutions, including the absence of structure for identified topics, non-standard aspect names, and lack of abundant training data. 
The proposed solution builds a semi-supervised multi-level taxonomy from raw reviews, a semantic similarity heuristic approach to generate labelled data and employs a multi-task insight extraction architecture by fine-tuning an LLM.
InsightNet identifies granular actionable topics with customer sentiments and verbatim for each topic. 
Evaluations on real-world customer review data show that InsightNet performs better than existing solutions in terms of structure, hierarchy and completeness. 
We empirically demonstrate that InsightNet outperforms the current state-of-the-art methods in multi-label topic classification, achieving an F1 score of 0.85, which is an improvement of 11\% F1-score over the previous best results.
Additionally, InsightNet generalises well for unseen aspects and suggests new topics to be added to the taxonomy.

\end{abstract}

\vspace{-0.2cm}
\section{Introduction}
\vspace{-0.15cm}

Customer reviews provide rich insights for various stakeholders, such as businesses, brands, and customers. They can inform product development, enhance customer experience, track reputation, and guide purchase decisions. However, customer reviews pose several challenges for analysis, such as subjectivity, variation, noise, domain-specificity, volume, and dynamism. Existing solutions for extracting structured insights from reviews, such as topic classification~\cite{zheng2021classification, sanchez2019naive}, polarity identification~\cite{bilal2022effectiveness, gopi2023classification}, and verbatim extraction~\cite{majumder2022perceived}, suffer from several drawbacks that limit their effectiveness and applicability. These drawbacks include: (1) low accuracy and reliability in generating and extracting insights from reviews, (2) lack of coherence and clarity in the output, which makes it hard to act upon, (3) high dependency on large amounts of annotated data, which are scarce and expensive to obtain, (4) task-specificity, (5) inability to handle multiple tasks simultaneously, (5) skewed data distribution towards a few dominant aspects, which biases the models' performance (81\% of reviews are covered by just 12\% of topics, see Appendix~\ref{appendix-sec:heavy-tailed-distrubution} for detailed analysis), (6) reliance on predefined aspects and limitations to discover new topics.

In this paper, we present three key modules to address the challenges of the existing approaches as follows:
(1) AutoTaxonomy: A method to generate a hierarchical taxonomy of aspects with minimal supervision (section~\ref{sec:taxonomy}). This helps to organise the output in a structured and hierarchical form 
(2) SegmentNet: An unsupervised data creation technique using semantic similarity based heuristics to produce labelled data (section~\ref{sec:segmentNet}) that contains topic, polarity and verbatim for each review. Here, verbatim is the exact segment of the review that describes the topic identified. 
(3) InsightNet: A generative model for insights extraction. We model aspect identification as a multi-task hierarchical classification problem and then leverage the generative model (section~\ref{sec:architecture}) to classify topic (granular aspect), identify sentiment, extract verbatim and also discover new topics that are not in the current taxonomy. We use T5-base~\cite{raffel2020exploring} as a pre-trained Large Language Model (LLM) and fine-tune it with the data obtained from SegmentNet. Thus, we do not require any manually annotated data to train InsightNet.

\vspace{-0.2cm}
\section{Related Work}
\vspace{-0.2cm}
\begin{figure*}
\includegraphics[width=\textwidth]{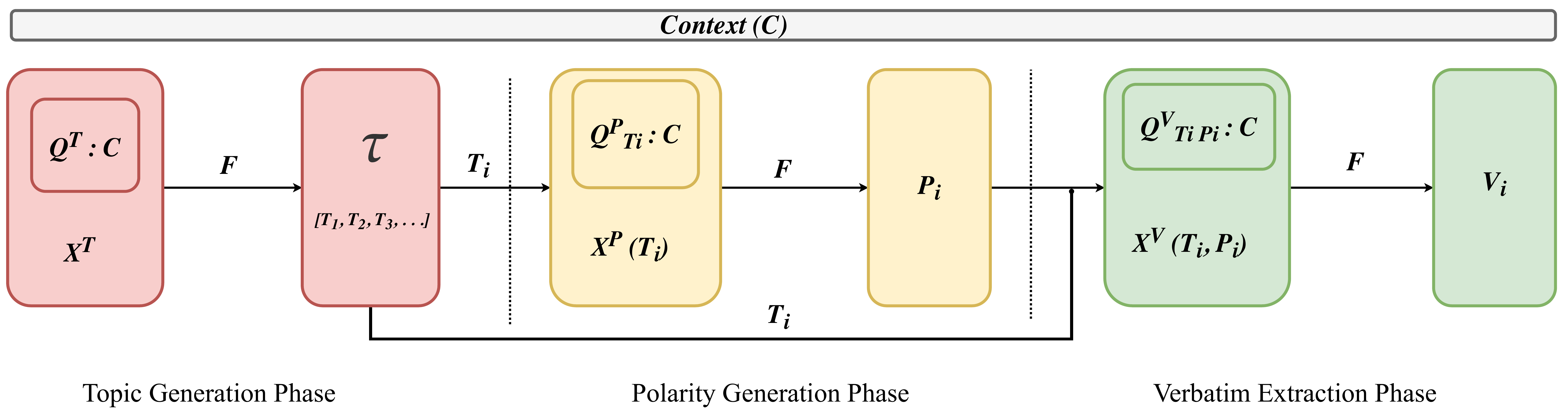}
\caption{Decomposed Sequential Prompting - InsightNet}
\label{fig:insightnet_seq_prompt_architecture}
\end{figure*}

Insight extraction from customer reviews is a well researched problem. Researchers have posed this problem in various frameworks such as heuristic based insight extraction, aspect based sentiment analysis, text summarisation, topic modeling, generative modeling. \citet{rana2015hybrid,kang2017rube} proposed a rule-based approach to extract insights from reviews. However, these approaches require huge manual efforts and domain expertise to discover patterns, update them frequently as new products launch and create rules. \citet{hu2004mining, baccianella2009multi} propose aspect based sentiment analysis methods which first extract aspects and then rate reviews on each aspect. However, aspects obtained using these methods are not granular enough for actionability. \citet{titov2008modeling, brody2010unsupervised, sircar2022distantly} proposed unsupervised approaches for aspect and sentiment analysis from reviews, but these methods suffer from two main limitations : a) redundancy of clusters and b) low interpretability, as the clusters produced are not actionable, structured or intuitive. Recently, generative approaches~\cite{raffel2020exploring, brown2020language} demonstrated promising performance on wide range of Natural Language Processing (NLP) tasks. \citet{liu2022leveraging} used a seq-to-seq model to generate product defects and issues from customer reviews, but they lack structure.

\vspace{-0.2cm}
\section{Problem Statement}
\vspace{-0.2cm}
Given a customer review $C$, we aim to extract a set of all the relevant and actionable insights $I_1, I_2, \ldots, I_k$, where each insight $I_i$ is composed of a granular topic $T_i$, a corresponding polarity $P_i$, and a set of verbatims $V_i$ associated with it.
\vspace{-0.2cm}
\section{Methodology}
\vspace{-0.2cm}
In this section, we present our generative approach, InsightNet, for mining insights (topics, polarities, verbatims) from raw reviews (obtanied from Amazon US marketplace\footnote{https://huggingface.co/datasets/amazon\_us\_reviews \label{foot-ref:amazon_us_reviews}}). Next, we describe how we construct a multi-level hierarchical taxonomy from the reviews to help organize topics in a meaningful way. Then, we introduce how the labelled data is created using SegmentNet. Later, we explain how we apply post-processing techniques to eliminate redundant topics and surface new topics that are not covered by the taxonomy. Finally, we discuss the experiments that led us to the design of InsightNet.
\subsection{InsightNet: Generative Multi-task model for Insights Extraction}
\label{sec:architecture}
\vspace{-0.1cm}
The InsightNet architecture (Figure~\ref{fig:insightnet_seq_prompt_architecture}) is based on decomposed prompting~\cite{khot2023decomposed}, allowing to solve the complex task of extracting actionable verbatims and assigning a topic and a polarity to each verbatim. It consists of three phases of prompting, one for topic generation, one for polarity generation, and one for verbatim extraction.

\subsubsection{Topic Generation Phase}
\label{sec:topic_generation}
We construct prompt $X^T$ by appending question $Q^T$ to context $C$ (raw review), where $Q^T$ is question prompt to generate list of granular actionable topics $\tau : [T_1, T_2, T_3, ...]$. We feed InsightNet model $(F)$, with $X^T$ to generate actionable topics list, $\tau$ 
\vspace{-0.5cm}
\begin{equation} X^T = Q^T : C \quad ;  \quad \tau = F(X^T) \end{equation}
\subsubsection{Polarity Generation Phase}
\label{sec:polarity_generation}
In this phase, we use the model $(F)$ sequentially to generate the polarity $(P_i)$ for each of the topic $(T_i)$ extracted in the previous phase. We feed $X^P(T_i)$, which consists of the context $C$ and the question prompt $Q^P_{T_i}$ for the topic $T_i \in \, \tau$. We then form $\Pi$, a set of topic-polarity pairs, $\Pi:[(T_1, P_1), (T_2, P_2), (T_3,P_3),...]$
\vspace{-0.25cm}
\begin{equation}
    X^P(T_i)=Q^P_{T_i}:C
\quad ;  \quad
    P_i = F(X^P(T_i))
\end{equation}
\subsubsection{Verbatim Extraction Phase}
\label{sec:verbatim_extraction}
In this last phase, we use the model $(F)$ sequentially to extract  verbatim $(V_i)$ for each topic-polarity pair $(T_i, P_i)$ produced previously. We feed  $X^V(T_i, P_i)$ to the model, which consists of the context  $C$ and the question prompt $Q^V_{T_i,P_i}$ for the pair $(T_i,P_i) \in \, \Pi$.
\vspace{-0.3cm}
\begin{multline}
    X^V(T_i,P_i) = Q^V_{T_i,P_i} : C  \quad ;  \\
    V_i = F(X^V(T_i, P_i))
\end{multline}
For a review, if we get $N$ topics in the first stage, then we subsequently use $N$ prompts for each of the next two stages. Thus, we use a total of $2N+1$ prompts per review, where $N$ is the number of actionable topics present in the review.
\begin{figure}[H]
\centering
\includegraphics[width=0.8\columnwidth]{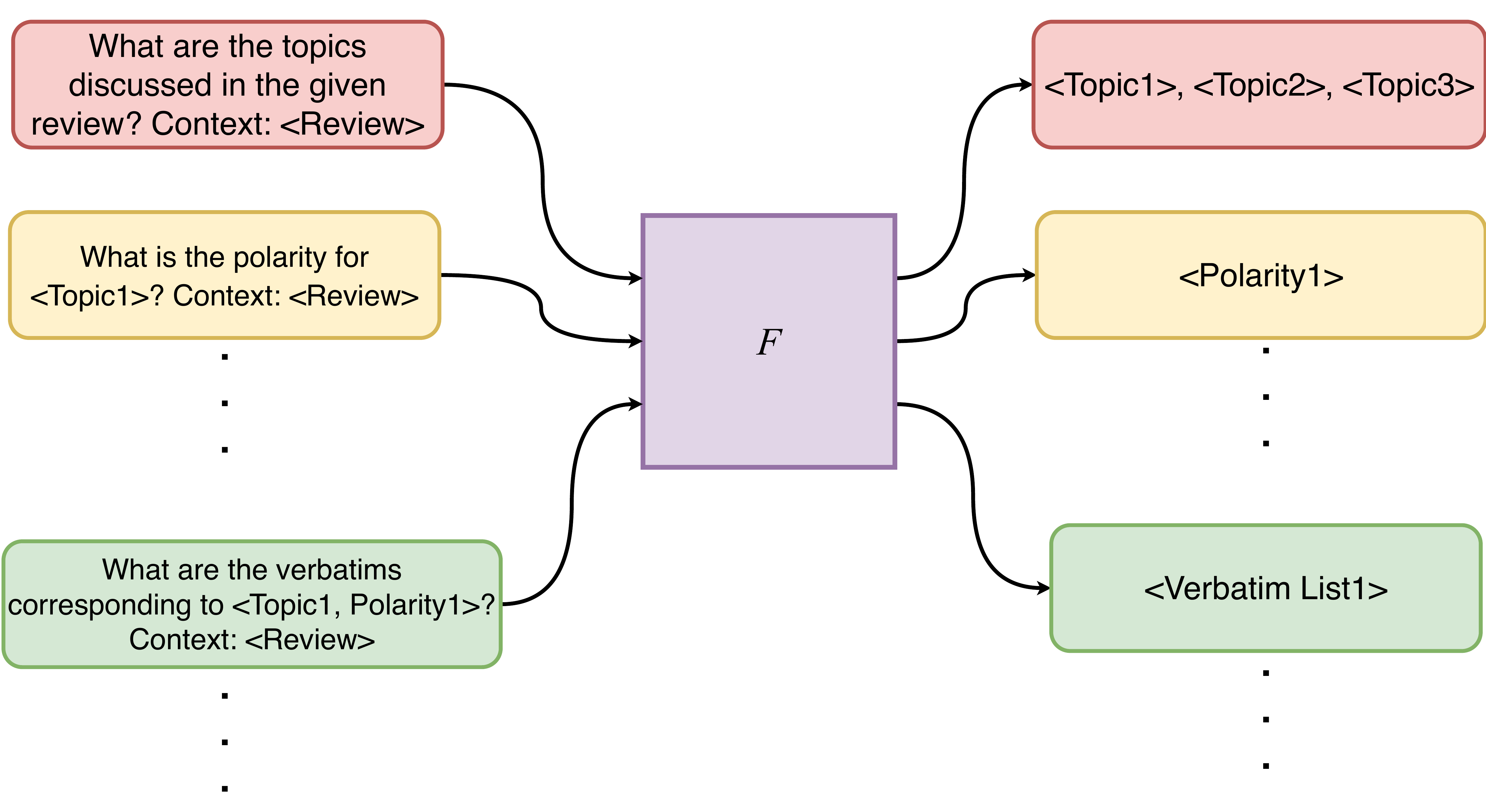}
\caption{InsightNet Prompting}
\label{fig:insightnet_prompts}
\end{figure}

\vspace{-0.5cm}
\subsection{AutoTaxonomy: Semi-supervised Taxonomy Creation}
\label{sec:taxonomy}
We propose a bottom-up method to generate a hierarchical auto-taxonomy from reviews with weak supervision. This means we start with identifying Granular Topics from the reviews, then group them into broader (high-level) topics. This helps us preserve structure and create hierarchy for the output. We segment raw reviews (refer to Appendix section~\ref{appendix-sec:segmentation_heuristics} for exact segmentation steps) and assign a polarity to each segment (see equation~\ref{eq:segmentnet_sentiment_classification}). We discard segments with neutral polarity. The following steps illustrate the process of Taxonomy creation:
 captured in 
1. \textbf{Clustering review segments}: We cluster the positive and negative segments separately using Fast clustering\footnote{\url{https://github.com/UKPLab/sentence-transformers/blob/master/examples/applications/clustering/fast_clustering.py}}, a sentence transformer~\cite{reimers-2019-sentence-bert} method. It uses cosine similarity to cluster sentence embeddings based on a threshold value. We obtain clusters for each sentiment class, representing different aspects or topics that the reviewers mentioned in their feedback.

2. \textbf{Merging similar clusters \& Cluster naming}: Human annotators merge duplicate clusters and name each cluster using pre-defined taxonomy guidelines. They name each cluster with fine-grained topic names that reflect the main idea of that cluster, forming the Granular Topics (Level-3) of the taxonomy.

3. \textbf{Creating hierarchy}: To structure the taxonomy, we group similar Granular Topics into Hinge Topics (Level 2) and Coarse Topics (Level 1), resulting in a multi-level hierarchical taxonomy.

4. \textbf{Keyword generation}: To get an exclusive and exhaustive set of keywords, we refined the clusters (of segments) obtained in step 2 (above). We applied the semantic similarity function (equation~\ref{eq:simst}) to perform \textbf{Intra-cluster} (refer section~\ref{sec:intra-cluster-cleaning}) cleaning to remove redundant and semantically duplicated keywords and \textbf{Inter-cluster} (refer section~\ref{sec:inter-cluster-cleaning}) cleaning of keywords, to eliminate ambiguous and overlapping keywords.

\subsubsection{Intra-cluster Cleaning}
\label{sec:intra-cluster-cleaning}
Given a set of keywords $K = \{k_1, k_2, \dots, k_n\}$, Algorithm~\ref{alg:iacc} returns a cleaned set of non redundant keywords $C$ such that $\forall k_i, k_j \in C$, $\textit{SimST}(k_i, k_j) \leq \delta_a$ 
\begin{algorithm}[H]
\caption{Intra-cluster cleaning}
\label{alg:iacc} 
\begin{algorithmic}[1]
\State Let $K = \{k_1, k_2, \dots, k_n\}$ are the set of keywords of a given topic $T$
\State Let $\delta_a$ be the intra-cluster threshold for redundancy 
\State Initialize $D \gets \emptyset$, $C \gets K$ \ForAll{$(k_i, k_j) \in K \times K$} \If{$\textit{SimST}(k_i, k_j) > \delta_a$} \State Add $k_i$ or $k_j$ to $D$ \EndIf \EndFor 
\State Remove all elements in $D$ from $C$ 
\State Return $C$ 
\end{algorithmic} 
\end{algorithm}

\subsubsection{Inter-cluster Cleaning}
\label{sec:inter-cluster-cleaning}
Algorithm~\ref{alg:iecc} compares the keywords across all the topics and removes keywords which are similar to each other by converting the keywords into sentence embedding and comparing the cosine similarity between them with the ambiguity threshold $\delta_e$.
\begin{algorithm}[t]
\caption{Inter-cluster cleaning} 
\label{alg:iecc} 

\begin{algorithmic}[1]
\State Let $K = {K_1, K_2, \dots, K_n}$ be the initial set of keyword lists respectively for $n$ topics $T = {T_1, T_2, \dots, T_n}$. 
\State Let $\delta_e$ be the inter-cluster threshold for ambiguity. 
\State Create an empty hash table $H$. 
\For {each keyword list $K_i$ in $K$} 
\For {each keyword $k_{ij}$ in $K_i$} 
\If {$k_{ij}$ is not in $H$} 
\State Compute $\textit{Sbert}(k_{ij})$ and store it in $H$ with $k_{ij}$ as the key and $\textit{Sbert}(k_{ij})$ as the value. 
\EndIf 
\EndFor 
\EndFor 

\For {each keyword list $K_i$ in $K$} 
\For {each keyword $k_{ij}$ in $K_i$} 
\For {each other keyword list $K_l$ in $K$ where $l \neq i$} 
\For {each keyword $k_{lm}$ in $K_l$} 
\If {$\textit{SimST}(H[k_{ij}], H[k_{lm}]) > \delta_e$} 
\State Remove $k_{ij}$ from $K_i$ and $k_{lm}$ from $K_l$ 
\EndIf 
\EndFor 
\EndFor 
\EndFor 
\EndFor 
\State Return the final keyword lists $K$. 
\end{algorithmic} 
\end{algorithm}

Taxonomy derived using this approach has 91\% exclusivity (uniqueness of topics) and 94\% exhaustivity of topics when evaluated manually. (Table~\ref{tab:sample_taxonomy} in Appendix section~\ref{appendix-sec:sample_taxonomy} presents sample of final taxonomy).\vspace{-0.2cm}


\subsection{SegmentNet: Data Generation Mechanism }
\label{sec:segmentNet}

\begin{algorithm}[bht!]
\caption{Topic matching}
\label{alg:topicmatching} 
\begin{algorithmic}[1]

\State // Returns the most relevant topic $T$ for a given segment $S$ by applying heuristic rules.
\State Hyper-parameters: $k = 5, \delta_h = 0.8, \delta_m = 0.3, \delta_a = 0.5$
\State $ T_{\textrm{n}}, S_{\textrm{n}} = BestTopicAndScore([T_i]_{i=1}^{N}),$
$[ SimST(S, T_i) ]_{i=1}^{N} )$
\State $ T_{\textrm{tkw}}, S_{\textrm{tkw}} = BestTopicAndScore([T_i]_{i=1}^{N},$ 
$[ \frac{1}{k} max_{p} \{ SimST(S, k_{i,j}) \}_{j=1}^{M_i} ]_{i=1}^{N})$

\State $ T_{\textrm{mkw}}, S_{\textrm{mkw}} = BestTopicAndScore$
$([T_i]_{i=1}^{N}, [ \frac{1}{M_i} \sum_{j=1}^{M_i} \{ SimST(S, k_{i,j}) \} ]_{i=1}^{N})$

\State $A_i = simST(S, T_i) + $
$\frac{1}{k} max_{p} \{ SimST(S, k_{i,j}) \}_{j=1}^{M_i} +$
$\frac{1}{M_i} \sum_{j=1}^{M_i} \{ SimST(S, k_{i,j}) \}$
\State $ T_{\textrm{avg}}, S_{\textrm{avg}} = BestTopicAndScore$
$(\{A_i\}_{i=1}^{N})$
\State

\If{$ S_{\textrm{tkw}} \geq \delta_h $} 
    \State $ T = T_{\textrm{tkw}} $
\ElsIf{$ S_{\textrm{n}} \geq \delta_h $} 
    \State $ T = T_{\textrm{n}} $
\ElsIf{$ S_{\textrm{mkw}} \geq \delta_h $} 
    \State $ T = T_{\textrm{mkw}} $
\ElsIf{$ T_{\textrm{tkw}} = T_{\textrm{n}} $ \textbf{and} $ S_{tkw} + S_{n} \geq 2*\delta_m $
    \State $ T = T_{\textrm{tkw}} $}
\ElsIf{$ T_{\textrm{mkw}} = T_{\textrm{tkw}} $ \textbf{and} $ S_{mkw} + S_{tkw} \geq 2*\delta_m $
    \State $ T = T_{\textrm{mkw}} $}
\ElsIf{$ T_{\textrm{n}} = T_{\textrm{mkw}} $ \textbf{and} $ S_{n} + S_{mkw} \geq 2*\delta_m $}
    \State $ T = T_{\textrm{n}} $
\ElsIf{$ S_{\textrm{avg}} \geq \delta_a $}
    \State $ T = T_{\textrm{avg}} $
\Else
    \State $ T = \emptyset $
\EndIf
\State
\State \textbf{return} T
\end{algorithmic} 
\end{algorithm}
SegmentNet is a semantic matching algorithm that generates high quality training data with minimal training. It extracts insights from reviews using the taxonomy alone. It assumes that insights are often in short phrases within a review. It produces insights at a segment level and then aggregates them at review level. This involves 3 major steps:

1.\textbf{Segmentation}: We use language syntax heuristics to split a review into segments. We observe that a segment typically has one sentiment and at most one topic.

2. \textbf{Sentiment classification}: We train a BERT-based model with two linear heads (one for +ve and one for -ve) to get the sentiment of segment 
\vspace{-0.3cm}
\begin{equation}
    p,\, n = SentimentClassifier(S)
\label{eq:segmentnet_sentiment_classification}
\end{equation}
We use $\delta_p = 0.7$ as the classification threshold. A segment is neutral if $p < \delta_p$ and $n < \delta_p$, where $p$ and $n$ are the probabilities of a verbatim being positive and negative polarities respectively. We fine-tune sentiment classification model on $80k$ segments with almost equal data for each label, which has 99.1\% accuracy when evaluated manually. \newline
\textbf{\textit{SimST}}: We formulate semantic similarity function SimST (equation \ref{eq:simst}) between two texts $text_i$ and $text_j$, where \textit{sbert} computes the Sentence-BERT~\cite{reimers-2019-sentence-bert} embedding of text. 
\vspace{-0.5cm}
\begin{multline} 
SimST(text_{i}, text_{j}) = \cos (\, sbert(text_{i}), \, \\ sbert(text_{j})) \label{eq:simst} 
\vspace{-0.2cm}
\end{multline} where $\cos(\mathbf{u}, \mathbf{v}) = \frac{\mathbf{u} \cdot \mathbf{v}}{|\mathbf{u}| |\mathbf{v}|}$ is the cosine similarity.

\begin{figure*}[t!]
\includegraphics[width=\textwidth]{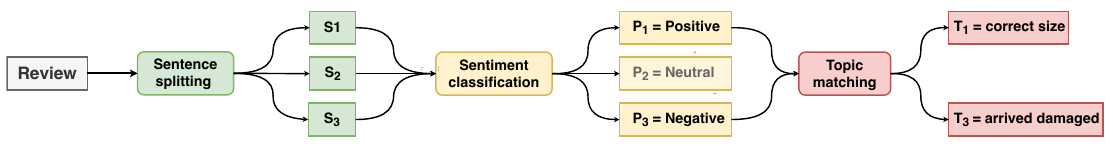}
\caption{SegmentNet pipeline}
\label{fig:segmentnet_pipeline}
\end{figure*}

\begin{algorithm}[H]
\caption{Signalling Algorithm ($BTS$)}
\label{alg:besttopic} 
\begin{algorithmic}[1]
\Procedure{BestTopicAndScore}{T, X}
    \State // Finds the leading topic $T_i$ as per the score values mentioned in the list $X$.
    \State \textbf{return} T[$argmax(X)$], $max(X)$
\EndProcedure
\end{algorithmic} 
\end{algorithm}

3. \textbf{Topic matching}: We devised heuristics based on the semantic matching function SimST (equation~\ref{eq:simst}) and a signalling algorithm (see BTS Algorithm~\ref{alg:besttopic}) to assign the best matching topic to a segment from a list of taxonomy topics. The signalling algorithm outputs the topic with the maximum similarity score and the value of that score among the given topic and similarity score pairs. Let $S$ denote a segment and $T$ its most relevant topic. We find $T$ from the list of taxonomy topics ($\tau'$), $[ T_i ]_{i=1}^{N}$ with each topic $T_i$ has keywords $[ k_{i,j} ]_{j=1}^{M_{i}}$. We define three signals (using Algorithm~\ref{alg:besttopic}), where the first signal (equation~\ref{signal:sematically_closest_topic}) is the semantically closest topic name and its score, the second signal (equation~\ref{signal:topk_best_mean_score_topic}) is the topic with best mean score with the five closest keywords, and the last signal (equation~\ref{signal:all_best_mean_score}) is the topic with the best mean score with all keywords. 
\begin{multline}
    T_{n}, \, S_{n} = BestTopicAndScore([T_i]_{i=1}^{N}), \\
    [SimST(S, T_i)]_{i=1}^{N})
\label{signal:sematically_closest_topic}
\end{multline}
\vspace{-0.4cm}
\begin{multline}
    T_{tkw}, \, S_{tkw} = BestTopicAndScore([T_i]_{i=1}^{N}, \\
    [\frac{1}{5} max_{5} \{ SimST(S, k_{i,j}) \}_{j=1}^{M_i} ]_{i=1}^{N})
\label{signal:topk_best_mean_score_topic}
\end{multline}
\vspace{-0.4cm}
\begin{multline}
    T_{mkw}, \, S_{mkw} = BestTopicAndScore([T_i]_{i=1}^{N}, \\
    [\frac{1}{M_i} \sum_{j=1}^{M_i} \{ SimST(S, k_{i,j}) \} ]_{i=1}^{N})
\label{signal:all_best_mean_score}
\end{multline}
To identify most relevant topic, we use heuristics on the three signals for topic matching: 

(a.) \textbf{High confidence match}: if any of the three signal scores is high, match with high scoring topic ($score \geq \delta_h$). This matches a segment that is very similar to a topic or keyword,

(b.) \textbf{Majority vote}: If any two signals give the same topic, match with the common topic ($score \geq \delta_m$). Since each of the three signals is an independent weak predictor of the correct topic, the fact that any two signals agree on a topic is a strong indicator of correctness,

(c.) \textbf{Best average score}: Match with the topic with the best average score across all three signals $T_{avg}$ \ ($score \geq \delta_a$).

We present the topic matching algorithm (Algorithm~\ref{alg:topicmatching}) which is more robust to noisy keywords and identifies topics with higher precision than simple semantic matching.

\subsection{Post-Processing}
\label{sec:post_processing}
During inference, we leverage syntactic and semantic matching to tackle topics generated that are out-of-taxonomy and re redundant. We either enrich taxonomy with these topics as fine-grained subtopics (L4 topics) or as novel topics (new L3 topics).
\subsubsection{Syntactic Matching}
\label{sec:syntactic_matching}
Let $gT$ be the generated topic and $\tau'$ be the set of topics in the taxonomy.  We compare $gT$ with each topic in $\tau'$ for exact or partial match. If no match is found, we use semantic matching.
\vspace{-0.1cm}
\begin{equation}
gT \leftarrow
\begin{cases}
t & \text{if} \quad gT \, = \, t \,; \quad t \, \in \, \tau' \\
t & \text{if} \quad gT \subset t \, ; \quad t \in \, \tau' \\
gT & \text{otherwise} 
\end{cases}
\end{equation}
\vspace{-0.3cm}
\subsubsection{Semantic Matching}
\label{sec:semantic_matching}
We use a signalling algorithm (refer BTS Algorithm~\ref{alg:besttopic}) to compute the best matching topics, and corresponding scores for each of the generated topic and extracted verbatim. For each topic $T_i$ in the taxonomy topics list $\tau'$, we find the maximum similarity with the generated topic ($gT$) as: 
\vspace{-0.1cm}
\begin{multline} 
\textit{topic\textsubscript{t}}, \textit{score\textsubscript{t}} =  BestTopicAndScore([T_i]_{i=1}^{N}, \\
[\textit{SimST}(\textit{gT}, T\textsubscript{\textit{i}})]_{i=1}^{N})
\label{eq:topic_similarity}
\end{multline} 
\vspace{-0.1cm}
Similarly, for each verbatim $k_j$ in the set of verbatims $K_i$ for each topic $T_i$, we find the maximum similarity with the extracted verbatim ($eV$) as: 
\begin{multline} 
\textit{topic\textsubscript{v}}, \textit{score\textsubscript{v}} = BestTopicAndScore([T_i]_{i=1}^{N}, \\
[\max_{k \in K_i}(\textit{SimST}(\textit{eV}, k))]_{i=1}^{N})
\label{eq:verbatim_similarity}
\end{multline} 
We use the above scores and a semantic post-processing heuristics (refer Algorithm~\ref{alg:semsim}) to mark the generated topic as a new topic (new L3), a fine-grained subtopic (L4) of an existing L3 topic, or an existing L3 topic. 

\begin{algorithm}
\caption{Semantic Matching} 
\label{alg:semsim}
\begin{algorithmic}[1]
\Require $(topic_t, score_t), (topic_v, score_v) $
\If{$score_t > 0.95$} 
\State replace generated\_topic with taxonomy topic $topic_t $
\ElsIf{$score_t > 0.7$ and $score_v > 0.4$} 
\State surface the generated\_topic as new granular topic (L4) 
\Else 
\State surface as \textit{new\_topic} to be added to the taxonomy 
\EndIf 
\end{algorithmic} 
\end{algorithm}


\vspace{-0.2cm}
\section{Experiments}

\subsection{Data generation ablation}
\label{sec:data_generation_ablation}
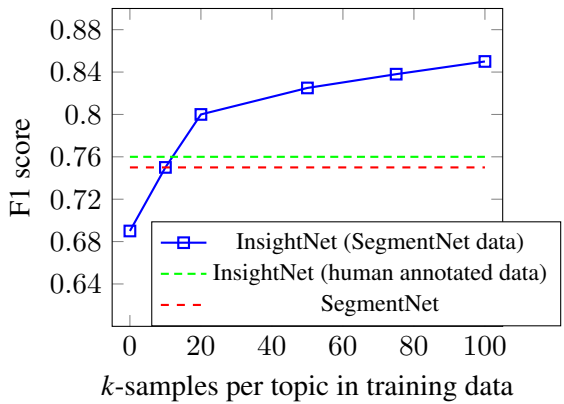
\begin{figure}[H]
\centering
\begin{tikzpicture}
\begin{axis}[
    width=0.42\textwidth,
    xlabel={\textit{k}-samples per topic in training data},
    ylabel={F1 score},
    xmin=-5, xmax=105,
    ymin=0.6, ymax=0.9,
    xtick={0, 20, 40, 60, 80, 100},
    ytick={0.64, 0.68,  0.72,  0.76,  0.8, 0.84, 0.88},
    legend style={at={(0.1,0)},anchor=south west},
    grid style=dashed,
]
\addplot[
    color=blue,
    mark=square,
    thick
    ]
    coordinates {
    {
        (0, 0.69)
        (10, 0.75)
        (20, 0.80)
        (50, 0.825)
        (75, 0.838)
        (100, 0.85)
    }
    };
    \addlegendentry{\footnotesize InsightNet (SegmentNet data)};
\addplot [green, densely dashed, thick ] coordinates {(0,0.76) (100,0.76)};
    \addlegendentry{\footnotesize InsightNet (human annotated data)};
\addplot [red, dashed, thick] coordinates {(0,0.75) (100,0.75)};
    \addlegendentry{\footnotesize SegmentNet};
\end{axis}
\end{tikzpicture}
\caption{SegmentNet Data Ablation}
\label{fig:sg_vs_in}
\end{figure}
\vspace{-0.2cm}
We show that SegmentNet can generate training data that is better or comparable to human annotated data. Figure~\ref{fig:sg_vs_in} compares the performance of InsightNet trained with SegmentNet on different dataset sizes with a \textit{fixed} human annotated dataset (fixed due to human bandwidth limitations). SegmentNet improves the performance by $9\%$ over human annotated data, given only $30$ samples/topic by manual annotation. This limitation is due to the heavy-tailed data (See Appendix section~\ref{appendix-sec:heavy-tailed-distrubution}) and the need for more data to cover the underrepresented topics. We also see that we need about three times more synthetic data to surpass the human-annotated baseline. We also show the model performance trained with \textit{k} samples per topic. We find that InsightNet outperforms SegmentNet around 20 samples per topic and stabilizes around 100 samples per topic.

\vspace{-0.25cm}
\subsection{Prompt Engineering}
\label{sec:prompt_engineering}
The choice of prompt can significantly affect the performance of language models like ours, especially in multi-task settings. We devised different variations of decomposed prompting for our multi-task problem of extracting actionable insights from customer reviews. We experimented with different orders of prompts for verbatim extraction ($V$), topic identification ($T_c$ and $T_g$), and polarity detection ($P$). We also explored different approaches for prompting for topic identification, either top-down (from coarse to granular) or bottom-up (from granular to coarse). We measured the performance of each variation using precision, recall, and F1-score metrics. We discovered that the optimal prompting strategy was to first prompt for the granular topics ($T_g$) from the review, then prompt for polarity ($P$) for each topic, and finally prompt for the verbatims ($V$) that correspond to the topics. This strategy achieved an F1-score of 0.80, which was considerably higher than the other variations. We also observed that using bottom-up prompting for topic identification was more efficient than using top-down prompting, as it minimized the errors in conditional prompting and enhanced the quality of topic extraction. We could deduce the coarse topics ($T_c$) from the granular topics ($T_g$) using the taxonomy. We refer to Level-1 and Level-2 topics as coarse topics and Level-3 topics as granular topics. We provide more detailed explanation of experiments in Appendix section~\ref{appendix-sec:prompt_engineering_discussion}.

\begin{table*}[ht!]
\centering
\renewcommand{\arraystretch}{1.1}
\resizebox{0.8\textwidth}{!}{
\begin{tabular}{lccccc}
\hline
               & \multicolumn{3}{c}{\textbf{\underline{Topic Classification (L3 + Polarity)}}}       & \multicolumn{2}{c}{\textbf{\underline{Verbatim Extraction}}} \\ 
\textbf{Model/Approach} & \textbf{Precision} & \textbf{Recall} & \textbf{F1 Score} & \textbf{Correctness}   & \textbf{Completeness}   \\ \hline
Multi Level Seq2seq \cite{liu2022leveraging}  & 0.34   & 0.38   & 0.36            & - & - \\
Rule-based \cite{rana2015hybrid} & 0.56   & 0.61   & 0.58            & - & - \\
BERT (ABSA)  \cite{hoang-etal-2019-aspect}         & 0.61 & 0.67 & 0.64          & - & - \\
DNNC - NLI \cite{zhang2020discriminative}             & 0.76 & 0.73 & 0.74          & - & - \\
Aspect Clustering \cite{sircar2022distantly}          & 0.70   & 0.79   & 0.74            & 0.70 & 0.97 \\
SegmentNet             & 0.82 & 0.70 & 0.76 & 0.82 & 0.98 \\
InsightNet             & \textbf{0.85} & \textbf{0.86} & \textbf{0.85} & \textbf{0.85} & \textbf{0.99} \\ \hline
\end{tabular}
}
\caption{InsightNet - Baselines \label{tab:insightnet_baselines}}

\end{table*}

\vspace{-0.2cm}
\subsection{AmaT5: Effect of Pre-Training}
\label{sec:pre_training}
We applied unsupervised pre-training \cite{li2021unsupervised} to fine-tune a pre-trained model with unlabeled data from the target domain to enhance its transferability. We used the T5-base model~\cite{raffel2020exploring} with review data (20M raw reviews) and the \textit{i.i.d. noise, replace spans} objectives to do this. We named the resulting model \textbf{AmaT5} and it showed better performance than the original. We also tried other variations, like T5-base, along with Sentence Shuffling (see section~\ref{sec:sentence_suffling}), BART~\cite{lewis2019bart}, FlanT5~\cite{chung2022scaling} and the results are shown in Table~\ref{tab:checkpoints}.

\begin{table}[H]
\centering
\resizebox{\columnwidth}{!}{
\begin{tabular}{lccc}
\hline
\textbf{LLM Checkpoint}          & \textbf{Precision} & \textbf{Recall} & \textbf{F1 Score} \\ \hline
T5-base                      & 0.79               & 0.81            & 0.80              \\
T5-base +\\

Sentence Shuffling & 0.80               & 0.81            & 0.80              
\\
BART                         & 0.71               & 0.74            & 0.72              \\
FlanT5                       & 0.81               & 0.83            & 0.82              \\
AmaT5                     & \textbf{0.85}      & \textbf{0.86}   & \textbf{0.85}     \\ \hline
\end{tabular}
}

\caption{InsightNet - Training Ablation} 
\label{tab:checkpoints}
\end{table}

\vspace{-0.5cm}
\subsection{Experimental Results \& Baselines}
\label{sec:existing_work_comparision}
We conducted a comprehensive evaluation of our proposed methodology across a diverse set of 43 categories, encompassing over 2200+ distinct product types, which collectively represent more than 95\% of the global volume of reviews. Our evaluation employed a dataset extracted from Amazon reviews\footref{foot-ref:amazon_us_reviews}. To ensure a robust assessment, both the training and test datasets were meticulously stratified at a granular topic level. Specifically, we used around $75k$ reviews for training and $10k$ reviews for testing, thereby ensuring coverage across all product categories and granular topics.

To facilitate an equitable comparison across different approaches, we carried out post-processing procedures, as outlined in Section \ref{sec:post_processing}. Our findings reveal that our approach outperforms Aspect Clustering \cite{sircar2022distantly}, a state-of-the-art method for topic extraction, in terms of coverage, diversity, and standardization. Specifically, our approach can generate over 1200+ unique topics that capture both positive and negative aspects of the reviews, while Aspect Clustering produces many redundant topics for the same level of coverage. Moreover, our approach ensures that the topics are consistent and coherent across reviews and product categories, with only 12\% of them being duplicates that can be easily merged in post-processing. On the other hand, Aspect Clustering approach faces some challenges in reducing high duplication rate and no standardization, meaning that many topics are redundant and suffer with duplicate entries.

Table \ref{tab:insightnet_baselines} shows the comparison of our results with the baselines. We also observed around 15\% new topics have emerged which were not part of taxonomy (detailed analysis in Appendix section~\ref{sec:new_topic_discovery}).

\vspace{-0.1cm}
\subsection{Why fine-tuning is required?} 
\label{sec:why_finetuning}
\vspace{-0.1cm}

In contrast to existing large language models like ChatGPT/OpenAI-GPT3~\cite{brown2020language}, Llama2/Meta~\cite{Touvron_Martin_2023}, Bard/Google-LaMDA~\cite{thoppilan2022lamda}, Falcon/TII~\cite{penedo2023refinedweb}, which fall short in extracting structured insights from customer reviews due to issues like generating redundant topics, domain-specificity, struggle to distinguish actionable vs non-actionable verbatims, and inefficiencies due large model size and inference latency. Additionally, they lack adherence to taxonomy topics as evidenced by our experiments (see Appendix section~\ref{appendix-sec:llm_without_finetuning} and Table~\ref{tab:llm-examples} for results), where only 7\% of reviews produced correct outputs, 11\% reasonable outputs, and the majority, 82\%, yielded random results.

\vspace{-0.2cm}
\section{Conclusion}
\vspace{-0.2cm}
We have presented InsightNet, a novel multi-task model that extracts granular insights from customer reviews. InsightNet jointly performs multi-topic identification, sentiment classification, and verbatim extraction for each review, generates new topics beyond existing taxonomy, and enriches taxonomy with consistent and exhaustive topics. InsightNet surpasses the state-of-the-art methods by 11\% F1-score on overall performance metrics, and achieves 85\% F1-score on topic classification. Furthermore, InsightNet is scalable and can handle various tasks with a structured and hierarchical output.
\bibliography{emnlp2023}
\bibliographystyle{acl_natbib}

\appendix

\section{InsightNet}

\subsection{Observations on usage of LLMs without fine-tuning}
\label{appendix-sec:llm_without_finetuning}
We experimented with different LLMs such as ChatGPT/OpenAI-GPT3, Llama-2/Meta, Bard/Google-LaMDA, Falcon/TII without any fine-tuning. We constructed the prompts using the review and the granular topics list and asked the LLM to predict the topic, polarity and verbatim for each review. The exact sequence of prompts used were: 
\begin{enumerate}
    \item \textbf{Topic generation}: \textit{Given the review <>, identify the topics discussed in the review from the list of topics (actionable aspects) in []},
    \item \textbf{Polarity generation}: \textit{Given the review <>, identify the polarity for each of these topics (actionable aspects) in []},
    \item \textbf{Verbatim extraction}: \textit{Given the review <>, extract the verbatim (review segment) corresponding to the topic-polarity list []}
\end{enumerate}
Table~\ref{tab:llm-examples} shows the predictions from different LLMs. It is evident that the pre-trained LLMs do not perform well on the specific tasks, even when given the Taxonomy topics as input. In most cases, the predicted topics are not part of the Taxonomy, and are substrings of the segments. Furthermore, the extracted verbatims are non-actionable as some of them have neutral polarity.

Hence, it is not recommended using them in production systems where the stakeholders expect structured and consistent outputs.

\subsection{Discussion on Prompt Engineering experiments}
\label{appendix-sec:prompt_engineering_discussion}
We tried different variations of decomposed prompting for our multi-task problem and arrived at final working prompts. We grouped the prompts into two categories:
\begin{enumerate}
\item  \textbf{Hierarchy of topic classification}:

   (a) \textbf{Top-down}: We prompted the model to infer a Coarse-grained Topic ($T_c$) from the review (R) first. Then, we used the review and the inferred $T_c$ as inputs to prompt the model to generate the corresponding Granular Topic ($T_g$).
   
   (b) \textbf{Bottom-up}: We prompted the model to generate a Granular Topic ($T_g$) from the review directly. We could derive the coarse topic ($T_c$) from $T_g$ using the existing taxonomy.
\item \textbf{Task ordering}: We also experimented with changing the order of the tasks. These are as follows:

    (a) Extracting actionable verbatim ($V$) first and then assigning topics to each verbatim
    
    (b) Generating Topic ($T_g$) first and then extracting verbatim for each topic
    
    (c) Extracting the  polarity ($P$) first followed by generating topics ($T_g$) for each polarity ($P$) followed by extracting verbatim ($V$)
\end{enumerate}
We used $PT_g$ to denote polarity specific granular topic extraction and $PT_c$ to denote polarity specific coarse topic extraction. We discussed the prompts, observations and conclusions for each experiment or prompting strategy in detail in Table \ref{tab:prompt_engineering}.

\begin{table*}
\centering
\begin{tabular}{p{3.5cm}cp{4.5cm}p{4cm}}
\textbf{Order} & \textbf{P/R/F1*} & \textbf{Observations} & \textbf{Next Steps} \\ \hline \hline
R $\rightarrow$ $V$ $\rightarrow$ $T_c$ $\rightarrow$ $T_g$  & 0.21/0.36/0.27 & The model could not differentiate between review segments that need action and those that do not, and extracted both types of verbatims. This caused wrong topic assignment to verbatims that are not actionable, leading to poor precision in topic identification. 
& 
To prevent this, first identify the topics and then extract the verbatims that match them. \\ \hline

R $\rightarrow$ $T_c$ $\rightarrow$ $T_g$  $\rightarrow$ $V$ & 0.34/0.38/0.36 & The model has difficulty in distinguishing the positive and negative aspects of the review. 
& 
Introduce a polarity-based topic identification prompt. \\ \hline

R $\rightarrow$($P T_c$ ) $\rightarrow$ $T_g$  $\rightarrow$ $V$ & 0.37/0.51/0.43 & The model often identifies topics that are contrary to the prompt’s polarity. 
& 
To avoid negative prompts generating positive topics and vice versa, first identify the topics and then assign the polarity.  \\ \hline

R $\rightarrow$ $T_c$  $\rightarrow$ $T_g$ $\rightarrow$ $P$ $\rightarrow$ $V$ & 0.42/0.53/0.47 & The accuracy of granular topic classification is affected by the low quality of coarse topic identification. This leads to errors in conditional prompting and poor metrics for granular topic identification. 
& 
Since the topics are organized in a hierarchical taxonomy, we can improve the results by starting with granular topic identification instead of coarse topic identification. Also, to reduce the number of prompts, we can directly prompt for polarity-based topic extraction. \\ \hline

R $\rightarrow$($P T_g$) $\rightarrow$ $V$ & 0.63/0.78/0.70 & The model identifies positive topics in negative prompts and negative topics in positive prompts. The model has difficulty in distinguishing between them and many of the topics are identified in both types of prompts, leading to poor precision. & To avoid this, first identify the actionable granular topics and then assign polarity to them. \\ \hline

R $\rightarrow$ $T_g$ $\rightarrow$ $P$ $\rightarrow$ $V$ & 0.79/0.81/0.80 & The model performed well in all three tasks. & Using the taxonomy, we can infer the higher levels: Coarse topics ($T_c$) and Hinge topics ($T_h$) \\ \hline
\end{tabular}

\caption{Prompt engineering approaches. *Note: The topics are post-processed (refer section \ref{sec:post_processing}) to match the taxonomy topics before calculating the metrics. The metrics are for the Granular Topic classification task.\label{tab:prompt_engineering}}
\end{table*}

\subsection{Observations on new topic discovery}
\label{sec:new_topic_discovery}

We analyzed $\sim10k$ reviews spanning across product categories and found that our model generated $\sim$1450+ unique topics. Out of these, $\sim$1200+ topics matched the existing taxonomy, while $\sim$200+ topics ($\sim$20\%) were new and emerged from post-processing. From the new topics discovered about $\sim$15\% of them were fined-grained subtopics (surfaced as L4 topics) versions of the existing granular (L3) topics, and the rest were completely new topics (surfaced as new L3) identified by InsightNet, which were not present in the base taxonomy. Note: We are not revealing exact numbers to comply with company legal policy.

\subsection{Post Processing Heuristics}
\label{appendix-sec:post_processing}
To ensure the quality and structure of the taxonomy, the post processing heuristics evaluates the scores of the generated topics. It then determines: (1) if a topic is new, or (2) if it can be a more granular topic (fine-grained subtopic L4) of an existing L3 topic, or (3) if it can be replaced by a similar topic from the taxonomy. This prevents the redundancy of topics that have the same meaning. Post processing also gives the hierarchical structure to the output.
This prevents the redundancy of topics that have the same meaning. Post processing also gives the hierarchical structure to the output.


\subsection{InsightNet Sample Predictions}
\label{appendix-sec:insightnet_sameple_predictions}
Table \ref{tab:insightnet-examples} shows the sample predictions from InsightNet.

\section{SegmentNet}
\label{appendix-sec:segmentnet}

\subsection{Segmentation Heuristics}
\label{appendix-sec:segmentation_heuristics}
We devised heuristics based on linguistic analysis which extracts meaningful phrases from reviews by splitting the text into sentences and then into phrases. Based on our analysis we fixed the minimum length of phrase to be 2 words to make the segment complete and meaningful.

\begin{enumerate}
    \item \textbf{Review $\rightarrow$  Sentences}: Split on \{ . ! ? ``but" \}
    \item \textbf{Sentence $\rightarrow$ \, Phrases}:
    \begin{itemize}
        \item Split sentence  on \{ , ; \& ``and" \}
        \item Do no split into phrases if any resulting phrases has $\leq$ 2 words
    \end{itemize}
\end{enumerate}

\section{Supplementary Material}
\subsection{Latency metrics}

Latency is measured on a $10MB$ dataset which contains about $20k$ reviews. 
Inference is done with a batch size of $32$ on a single $m5.12xlarge$ instance for CPU performance and $p3.2xlarge$ for GPU performance calculation.

\begin{table}[H]
\centering
\caption{Infernence latency comparison}
\begin{tabular}{ccc}
\hline
        & \textbf{InsightNet}   & \textbf{SegmentNet}   \\ \hline
    CPU & 5hrs 33mins           & 5hrs 40 mins          \\
    GPU & 2hrs 30mins           & 2hrs 40mins           \\
\hline
\end{tabular}
\label{tab:latency_metrics}
\end{table}

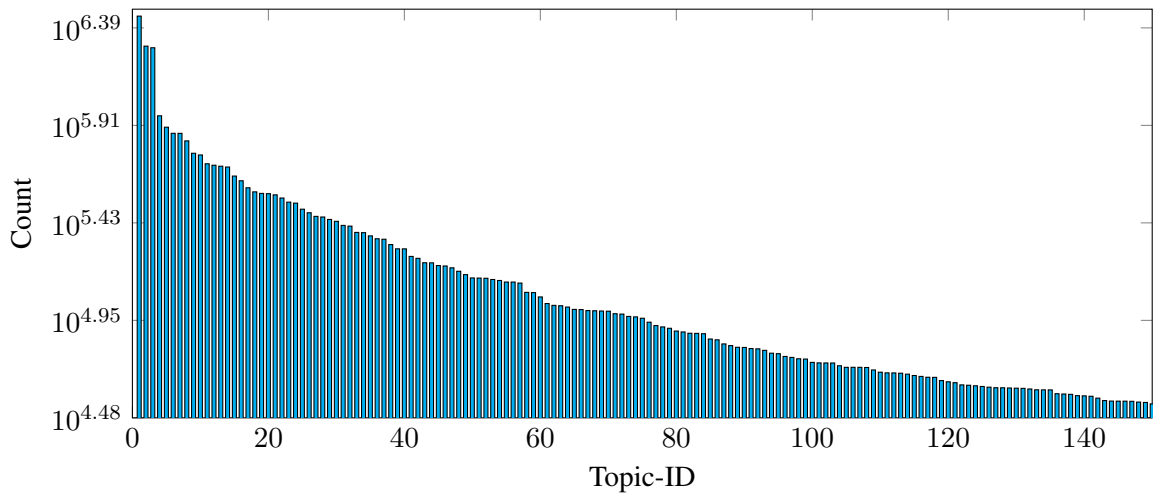
\begin{figure*}
\centering
\begin{tikzpicture}
\begin{axis}[
    width=15cm,
    height=7cm,
    xlabel={Topic-ID},
    ylabel={Count},
    ymode=log,
    bar width=1.5pt,
    ytick distance=3,
  ymin=30000, ymax=3000000,
     xmin=0, xmax=150,
    legend pos=north east
]
\addplot[ybar,fill=cyan]
    coordinates {
    {
(1,2774796)
(2,1978779)
(3,1942751)
(4,902327)
(5,793716)
(6,740701)
(7,740565)
(8,680161)
(9,591597)
(10,580404)
(11,525460)
(12,516992)
(13,511260)
(14,506544)
(15,457555)
(16,433937)
(17,400995)
(18,383037)
(19,376386)
(20,375208)
(21,370259)
(22,357344)
(23,341133)
(24,337402)
(25,315056)
(26,302633)
(27,290531)
(28,288493)
(29,280632)
(30,274799)
(31,262407)
(32,260278)
(33,242492)
(34,241782)
(35,233188)
(36,225744)
(37,224590)
(38,211387)
(39,201537)
(40,201484)
(41,185043)
(42,181053)
(43,172282)
(44,172077)
(45,166944)
(46,166115)
(47,162717)
(48,156511)
(49,150647)
(50,145160)
(51,144964)
(52,144636)
(53,142655)
(54,141076)
(55,138614)
(56,138578)
(57,137010)
(58,123383)
(59,123074)
(60,117209)
(61,108765)
(62,106570)
(63,106121)
(64,104532)
(65,101850)
(66,101690)
(67,100486)
(68,100269)
(69,99998)
(70,99806)
(71,96985)
(72,96585)
(73,94095)
(74,93596)
(75,92159)
(76,88168)
(77,84874)
(78,83594)
(79,82316)
(80,79808)
(81,78989)
(82,77852)
(83,77750)
(84,77439)
(85,72947)
(86,72287)
(87,69026)
(88,67729)
(89,66447)
(90,66333)
(91,65487)
(92,65304)
(93,64194)
(94,62174)
(95,61825)
(96,59933)
(97,59281)
(98,58386)
(99,58195)
(100,56055)
(101,55749)
(102,55693)
(103,55673)
(104,53945)
(105,53033)
(106,53032)
(107,53027)
(108,52920)
(109,51465)
(110,50197)
(111,49815)
(112,49782)
(113,49621)
(114,49092)
(115,48349)
(116,47857)
(117,47408)
(118,47361)
(119,45658)
(120,45043)
(121,44587)
(122,43501)
(123,43335)
(124,43058)
(125,42733)
(126,42398)
(127,42130)
(128,42106)
(129,42006)
(130,41921)
(131,41833)
(132,41542)
(133,41160)
(134,41128)
(135,41111)
(136,39387)
(137,39255)
(138,39110)
(139,38525)
(140,38450)
(141,38274)
(142,37476)
(143,36453)
(144,36249)
(145,36221)
(146,36212)
(147,36149)
(148,35844)
(149,35701)
(150,35114)
}
    };

\end{axis}
\end{tikzpicture}
\caption{Heavy tailed distribution of topics}
\label{fig:topic_distribution}
\end{figure*}

\label{appendix-sec:heavy-tailed-distrubution}
The aspect mentions in the reviews are heavy tail distributed, since large number of review segments are dominated by minority number of topics. We have plotted the frequency or support for each of 1200+ topics for 6 month review data spanning across all product categories, and plotted the histogram and can be seen in Figure~\ref{fig:topic_distribution}.


\begin{table*}
\centering
\resizebox{2\columnwidth}{!}{
\begin{tabular}{p{2.5in}|p{3in}|p{3.5in}}
\hline
\textbf{Review} &
  \textbf{InsightNet - Model Prediction} &
  \textbf{InsightNet - Post Processed Output} \\ \hline \hline
Clear and wide angle view. Provides a clear view of a wide area. &
  {[}\{('visibility', 'positive'): {[}'Provides a clear view'{]}\}, 
  \newline \{('wide angle
  mode', 'positive'): {[}'Clear and wide angle view'{]}\}{]} &
 {[}\{'L1 Topic': 'specifications and functionality', 'L2 Topic': 'visibility', 'L3 Topic': 'visibility', 'Polarity': 'positive', 'Verbatims': {[}'Provides a clear view'{]}\}, 
  \newline \{'L1 Topic': 'specifications and functionality', 'L2 Topic': 'wide angle mode', 'L3 Topic': 'wide angle mode', 'Polarity': 'positive', 'Verbatims': {[}'Clear and wide angle view'{]}\}{]} \\ \hline
To much treble no bass. Sound was not good. To much treble no bass sounds tinny. &
  {[}\{('audio quality', 'negative'): {[}'Sound was not good'{]}\}, \newline \{('treble', 'negative'): {[}'To much treble no bass', 'To much treble no bass sounds tinny'{]}\}{]} &
  {[}\{'L1 Topic': 'specifications and functionality', 'L2 Topic': 'audio quality', 'L3 Topic': 'audio quality', 'Polarity': 'negative', 'Verbatims': {[}'Sound was not good'{]}\}, \newline \{'L1 Topic': 'specifications and functionality', 'L2 Topic': 'treble', 'L3 Topic': 'treble', 'Polarity': 'negative', 'Verbatims': {[}'To much treble no bass', 'To much treble no bass sounds tinny'{]}\}{]} \\ \hline
Short inseam. Fit was fine but theyre very short &
  {[}\{('in seam fit', 'negative'): {[}'Short inseam'{]}\}, \newline \{('correct fit', 'positive'): {[}'Fit was fine'{]}\}, \newline \{('size smaller than expected', 'negative'): {[}'theyre very short'{]}\}{]} &
  {[}\{'L1 Topic': 'design and make', 'L2 Topic': 'fit', 'L3 Topic': 'in seam fit', 'Polarity': 'negative', 'Verbatims': {[}'Short inseam'{]}\},  
  \newline
  \{'L1 Topic': 'design and make', 'L2 Topic': 'fit', 'L3 Topic': 'correct fit', 'Polarity': 'positive', 'Verbatims': {[}'Fit was fine'{]}\}, \newline \{'L1 Topic': 'design and make', 'L2 Topic': 'size', 'L3 Topic': 'size smaller than expected', 'Polarity': 'negative', 'Verbatims': {[}'theyre very short'{]}\}{]} \\ \hline
Not comfortable. The sole inside is very hard &
  {[}\{('comfort', 'negative'): {[}'Not comfortable'{]}\}, \newline \{('innersole quality', 'negative'): {[}'The sole inside is very hard'{]}\}{]} &
  {[}\{'L1 Topic': 'specifications and functionality', 'L2 Topic': 'ease of use comfort', 'L3 Topic': 'comfort', 'Polarity': 'negative', 'Verbatims': {[}'Not comfortable'{]}\}, \newline \{'L1 Topic': 'specifications and functionality', 'L2 Topic': 'material quality', 'L3 Topic': 'innersole quality', 'Polarity': 'negative', 'Verbatims': {[}'The sole inside is very hard'{]}\}{]} \\ \hline
No fishy after taste! I take this to help control cholesterol. &
  {[}\{('taste overall', 'positive'): {[}'No fishy after taste'{]}\}, \newline \{('heart health', 'positive'): {[}'I take this to help control cholesterol'{]}\}{]} &
  {[}\{'L1 Topic': 'specifications and functionality', 'L2 Topic': 'taste', 'L3 Topic': 'taste overall', 'Polarity': 'positive', 'Verbatims': {[}'No fishy after taste'{]}\}, \newline \{'L1 Topic': 'health and safety', 'L2 Topic': 'heart health', 'L3 Topic': 'heart health', 'Polarity': 'positive', 'Verbatims': {[}'I take this to help control cholesterol'{]}\}{]} \\ \hline
It's effective. This is probably the best magnesium supplement, have tried several. It's the best thus far in helping with sleep. &
  {[}\{('magnesium', 'positive'): {[}'This is probably the best magnesium supplement'{]}\}, \newline \{('sleep', 'positive'): {[}"It's the best thus far in helping with sleep"{]}\}{]} &
  {[}\{'L1 Topic': 'specifications and functionality', 'L2 Topic': 'nutritional value', 'L3 Topic': 'magnesium', 'Polarity': 'positive', 'Verbatims': {[}'This is probably the best magnesium supplement'{]}\}, \newline \{'L1 Topic': 'health and safety', 'L2 Topic': 'sleep quality', 'L3 Topic': 'sleep', 'Polarity': 'positive', 'Verbatims': {[}"It's the best thus far in helping with sleep"{]}\}{]} \\ \hline
Yes! Arrived nice and intact. No melting. No old taste. Good choice! &
  {[}\{('shipment or delivery', 'positive'): {[}'Arrived nice and intact'{]}\}, \newline \{('storage temperature', 'positive'): {[}'No melting'{]}\}, \newline \{('taste overall', 'positive'): {[}'No old taste'{]}\}{]} &
  {[}\{'L1 Topic': 'shipment package and delivery', 'L2 Topic': 'delivery experience', 'L3 Topic': 'shipment or delivery', 'Polarity': 'positive', 'Verbatims': {[}'Arrived nice and intact'{]}\}, \newline \{'L1 Topic': 'specifications and functionality', 'L2 Topic': 'storage temperature', 'L3 Topic': 'storage temperature', 'Polarity': 'positive', 'Verbatims': {[}'No melting'{]}\}, \newline \{'L1 Topic': 'specifications and functionality', 'L2 Topic': 'taste', 'L3 Topic': 'taste overall', 'Polarity': 'positive', 'Verbatims': {[}'No old taste'{]}\}{]} \\ \hline
Very pleased. I love this aloe. It's 100\% organic, feels very nice on skin. I'll use this for a variety of things. &
  {[}\{('organic', 'positive'): {[}"It's 100\% organic"{]}\}, \newline \{('skin health', 'positive'): {[}'feels very nice on skin'{]}\}{]} &
  {[}\{'L1 Topic': 'specifications and functionality', 'L2 Topic': 'nutritional value', 'L3 Topic': 'organic', 'Polarity': 'positive', 'Verbatims': {[}"It's 100\% organic"{]}\}, \newline \{'L1 Topic': 'health and safety', 'L2 Topic': 'health benefits', 'L3 Topic': 'skin health', 'Polarity': 'positive', 'Verbatims': {[}'feels very nice on skin'{]}\}{]} \\ \hline
Very sturdy and shreds well. Very sturdy solid machine. Can handle 3-4 pages at a time. Shreds into small bits!! &
  {[}\{('shredding quality', 'negative'): {[}'Shreds into small bits'{]}\}, \newline \{('strong', 'positive'): {[}'Very sturdy solid machine'{]}\}, \newline \{('paper quality', 'positive'): {[}'Can handle 3-4 pages at a time'{]}\}{]} &
  {[}\{'L1 Topic': 'specifications and functionality', 'L2 Topic': 'shredding quality', 'L3 Topic': 'shredding quality', 'Polarity': 'negative', 'Verbatims': {[}'Shreds into small bits'{]}\}, \newline \{'L1 Topic': 'specifications and functionality', 'L2 Topic': 'durability longevity', 'L3 Topic': 'strong', 'Polarity': 'positive', 'Verbatims': {[}'Very sturdy solid machine'{]}\}, \newline \{'L1 Topic': 'specifications and functionality', 'L2 Topic': 'material quality', 'L3 Topic': 'paper quality', 'Polarity': 'positive', 'Verbatims': {[}'Can handle 3-4 pages at a time'{]}\}{]} \\ \hline
\end{tabular}
}
\caption{Sample predictions from InsightNet \label{tab:insightnet-examples}}
\end{table*}

\begin{table*}
\centering
\resizebox{2\columnwidth}{!}{
\begin{tabular}{p{3.5in}|p{0.7in}|p{4in}}
\hline
\textbf{Review} &
\textbf{LLM} &
\textbf{Prediction} \\ 
\hline 
\hline
  
This has been a horrible experience for me.  It is a 5 cup coffee maker and I was attempting to make 4,it over flowed and has taken nearly 4 hours to brew one cup of coffee 
&
GPT-3 
&
\{ "topics": {[}"overflowed", "brew one cup of coffee"{]}, "verbatim": "It over flowed and has taken nearly 4 hours to brew one cup of coffee", "polarity": "Negative" \} 
\\ 
\hline

The opening of the bag is square, but the lid is non-existent;it's a zipper straight across the top (between two sides),which means the bag holds a lot less than it could. Disappointed.The bags were in their original flimsy plastic wrap. I don't recommend these bags. 
&
GPT-3 
&
\{  "topic": "opening of the bag",   "verbatim": "The opening of the bag is square, but the lid is non-existent; it's a zipper straight across the top (between two sides), which means the bag holds a lot less than it could.",   "polarity": "negative"\} 
\\ 
\hline

Very small. While they were cute, I had to give them to my 8 year old granddaughter. She's.the only one who could.wear them. 
&
Llama-2 &
\{"topics": {[}"age suitability", "fit",  "quality"{]}, "verbatim": "Very small. While they were cute, I had to give them to my 8 year old granddaughter. She's the only one who could wear them.", "polarity": "neutral"\}\} 
\\ 
\hline

Looks good, nice material, very tight in shoulders. I loved the material, both inside and outside. It’s comfortable, but somewhat loud (like a windbreaker). It’s very warm and lots of useful pockets, but it was way too tight in the shoulders to have range of motion. 
&
Llama-2 
&
\{"topics": {[}"material quality", "comfort", "warmth", "breathability", "range of motion"{]}, "verbatim": "Looks good, nice material, very tight in shoulders. I loved the material, both inside and outside. It’s comfortable, but somewhat loud"{]},
"polarities": {[} "positive", "positive", "neutral", "neutral", "negative" {]}
\} 
\\ 
\hline

Very small. While they were cute, I had to give them to my 8 year old granddaughter. She's.the only one who could.wear them. 
&
Bard 
&
\{ "review": "Very small. While they were cute, I had to give them to my 8 year old granddaughter. She's the only one who could wear them.", "topics": {[}"incorrect fit"{]}, "verbatim": "Very small. While they were cute, I had to give them to my 8 year old granddaughter. She's the only one who could wear them.", "polarity": "negative" \} 
\\ 
\hline

Looks good, nice material, very tight in shoulders. I loved the material, both inside and outside. It’s comfortable, but somewhat loud (like a windbreaker). It’s very warm and lots of useful pockets, but it was way too tight in the shoulders to have range of motion. 
&
Bard 
&
\{ "review": "Looks good, nice material, very tight in shoulders. I loved the material, both inside and outside. It’s comfortable, but somewhat loud (like a windbreaker). It’s very warm and lots of useful pockets, but it was way too tight in the shoulders to have range of motion","topics": {[}"fit", "material", "noise"{]}, "verbatims": {[} "very tight in shoulders", "loved the material, both inside and outside", "somewhat loud (like a windbreaker)" {]}, "polarities": {[} "negative", "positive", "neutral" {]} \} 
\\ 
\hline

This has been a horrible experience for me.  It is a 5 cup coffee maker and I was attempting to make 4,it over flowed and has taken nearly 4 hours to brew one cup of coffee 
&
Falcon 
&
\{ "topics": {[}"brew one cup of coffee"{]}, "verbatim": "It over flowed and has taken nearly 4 hours to brew one cup of coffee", "polarity": "negative" \} 
\\ 
\hline

The opening of the bag is square, but the lid is non-existent;it's a zipper straight across the top (between two sides),which means the bag holds a lot less than it could. Disappointed.The bags were in their original flimsy plastic wrap. I don't recommend these bags. 
&
Falcon 
&
\{"topic": square",   "verbatim": "The opening of the bag is square, but the lid is non-existent",   "polarity": "neutral"\} 
\\ 
\hline

\end{tabular}
}
\caption{Prediction using pre-trained LLMs}
\label{tab:llm-examples}
\end{table*}

\subsection{Discussion on Taxonomy}
\label{appendix-sec:sample_taxonomy}
Table~\ref{tab:sample_taxonomy} presents a sample of the hierarchical auto-taxonomy derived from reviews. The auto-taxonomy generated using reviews from 40+ product categories resulted in 8 L1 topics, 600+ L2 topics, and 1200+ L3 topics~\footnote{We are not revealing exact numbers to comply with company legal policy}.

\subsection{SegmentNet examples}
Table \ref{tab:segmentnet_examples} shows sample data generated using SegmentNet.
\begin{table*}
\centering
\scriptsize
\resizebox{2\columnwidth}{!}{
\begin{tabular}{l|l|l|l}
\hline
\textbf{Review} & \textbf{Segment} & \textbf{Polarity} & \textbf{Matched topic} \\ \hline
\hline
{\begin{tabular}[c]{@{}l@{}}\\ Not even close. \\ not even close to the same as the image.\end{tabular}} & Not even close & negative & no topic \\ \cline{2-4} 
 & \begin{tabular}[c]{@{}l@{}} not even close to \\ the same as the \\ image\end{tabular} & negative & false advertising \\ \hline
{\begin{tabular}[c]{@{}l@{}}\\ Color is GREAT! Have to battle the sleeve \\ tightness. Length is great. Warmth is there. \\ Just very tight in the arm area. \\ Not shoulders but sleeves\end{tabular}} & Color is GREAT! & positive & color \\ \cline{2-4}
 & \begin{tabular}[c]{@{}l@{}} Have to battle the sleeve \\ tightness \end{tabular} & negative & \begin{tabular}[c]{@{}l@{}}size smaller \\ than expected\end{tabular} \\ \cline{2-4} 
 & Length is great & positive & correct size \\ \cline{2-4} 
 & Warmth is there & positive & warmth \\ \cline{2-4} 
 & \begin{tabular}[c]{@{}l@{}}Just very tight in \\ the arm area\end{tabular} & negative & arm fit \\ \cline{2-4} 
 & not the shoulders & neutral & - \\ \cline{2-4} 
 & sleeves & neutral & - \\ \hline

\end{tabular}}
\caption{SegmentNet: Data Generation Examples}
\label{tab:segmentnet_examples}
\end{table*}

\subsection{Data Augmentation: Sentence Shuffling}
\label{sec:sentence_suffling}
Sentence shuffling is a data augmentation technique that we applied to the labelled reviews. We split each review into sentences based on full-stop and then randomly rearranged the sentences to form a shuffled review. The label of the shuffled review remained the same as the original review. We found that the average number of sentences in a review was between 3 and 6. Therefore, we could generate up to 6 shuffled versions of each review and add them to the training data to increase its size.

\begin{table*}
\centering
\begin{tabular}{p{2.7cm}|p{2.5cm}|p{2.2cm}|p{1.2cm}|p{3.5cm}}
\hline
\textbf{Coarse Topic} & \textbf{Hinge Topic} & \textbf{Granular Topic} & \textbf{Polarity} & \textbf{Keywords} \\ \hline \hline
customer service & responsiveness & great responsiveness & positive & replied fast, immediate response ... \\ \hline
customer service & responsiveness & unable to reach support & negative & no response, can't reach vendor ... \\ \hline
design and make & size & correct size & positive & size as expected, true to size ... \\ \hline
design and make & size & size larger than expected & negative & too long, bigger than expected ... \\ \hline
design and make & size & size smaller than expected & negative & too short, XXL fits like an L ... \\ \hline
health and safety & sleep quality & sleep quality & negative & not helpful for sleep, poor sleep assistance ... \\ \hline
health and safety & sleep quality & sleep quality & positive & sleep quality improves, good for active sleepers ... \\ \hline
specifications and functionality & material quality & zipper quality & negative & zipper sticks, does not zip well ... \\ \hline
specifications and functionality & material quality & zipper quality & positive & unzips smoothly, easy to zip ... \\ \hline
returns refunds and replacements & policies and initiation & cannot initiate returns & negative & no option to return, outside of return policy ... \\ \hline
returns refunds and replacements & policies and initiation & unclear policies & negative & no return policy, bad replacement policy ... \\ \hline
shipment package and delivery & packaging & good packaging & positive & safe and secure packaging, pleased with packaging quality ... \\ \hline
shipment package and delivery & packaging & package damaged & negative & box arrived crushed, package arrived with dents, envelope ripped open ... \\ \hline
shipment package and delivery & packaging & redundant packaging & negative & too much plastic in package, arrived with too many boxes, ... \\ \hline
shipment package and delivery & packaging & unhygienic packaging & negative & package has stains, arrived wet and soggy \\ \hline
used damaged expired & new used product & new product & positive & brand new product, condition is new ... \\ \hline
used damaged expired & new used product & refurbished product & negative & refurbished sent, it is clearly refurbished ... \\ \hline
used damaged expired & new used product & used product & negative & Has been used previously, was sent a used product ... \\ \hline
miscellaneous & advertising related & as advertised claimed & positive & works as advertised, specs match the description \\ \hline
miscellaneous & advertising related & false advertising & negative & pictures are deceiving, product different than expected \\ \hline
\end{tabular}
\caption{Sample Taxonomy \label{tab:sample_taxonomy}}
\end{table*}

\section{Discussion on Future work}
We intend to expand this approach to multilingual and multimodal settings. Furthermore, we plan to extend a model to perform additional tasks, such as summarising review insights at various levels of granularity.

\end{document}